\documentclass[10pt,twocolumn,letterpaper]{article}

\usepackage[pagenumbers]{cvpr} 

\usepackage{amsmath}   
\usepackage{bm}  

\usepackage{placeins}
\usepackage{float}
\usepackage{stfloats}
\definecolor{red}{rgb}{1,0.2,0.2}
\definecolor{or}{rgb}{1,0.5,0.25}
\definecolor{green}{rgb}{0, 1, 0}
\definecolor{bl}{rgb}{0, 0, 1}
\definecolor{brown}{rgb}{0.59, 0.3, 0}
\definecolor{cyan}{rgb}{0, 1, 1}
\definecolor{c_lowbest}{rgb}{1.0,1.0,0.9}
\definecolor{c_highbest}{rgb}{0.9,1.0,0.9}

\newcommand{\best}[1]  {\textcolor{red}{\textbf{#1}}}
\newcommand{\second}[1]  {\textcolor{blue}{\underline{#1}}}

\usepackage{multirow}

\newcommand*{\affaddr}[1]{#1} 
\newcommand*{\affmark}[1][*]{\textsuperscript{#1}}
\newcommand*{\email}[1]{\texttt{#1}} %

\definecolor{cvprblue}{rgb}{0.21,0.49,0.74}
\usepackage[pagebackref,breaklinks,colorlinks,allcolors=cvprblue]{hyperref}

\def\paperID{xxxxx} 
\def\confName{CVPR}
\def\confYear{2026}

\title{EcoSplat: Efficiency-controllable Feed-forward 3D Gaussian Splatting \\from Multi-view Images}

\author{
Jongmin Park\affmark[1]\footnotemark[1] \quad Minh-Quan Viet Bui\affmark[1]\footnotemark[1] \\ Juan Luis Gonzalez Bello\affmark[2]\quad Jaeho Moon\affmark[1] \quad Jihyong Oh\affmark[3]\footnotemark[2] \quad Munchurl Kim\affmark[1]\footnotemark[2]\\
\affaddr{\affmark[1]KAIST} \quad
\affaddr{\affmark[2]Flawless AI} \quad
\affaddr{\affmark[3]Department of Imaging Science, GSAIM, Chung-Ang University}\\
\footnotesize{\email{\{jm.park, bvmquan, jaeho.moon, mkimee\}@kaist.ac.kr}} \quad \footnotesize{\email{juanluisgb.phd@gmail.com}} \quad 
\footnotesize{\email{jihyongoh@cau.ac.kr}}\\
\affaddr{\small{\url{https://kaist-viclab.github.io/ecosplat-site/}}}%
}

\begin{document}
\twocolumn[{
\renewcommand\twocolumn[1][]{#1}%
\maketitle           
\vspace{-0.6cm}
\centering
\includegraphics[width=0.89\linewidth,keepaspectratio]{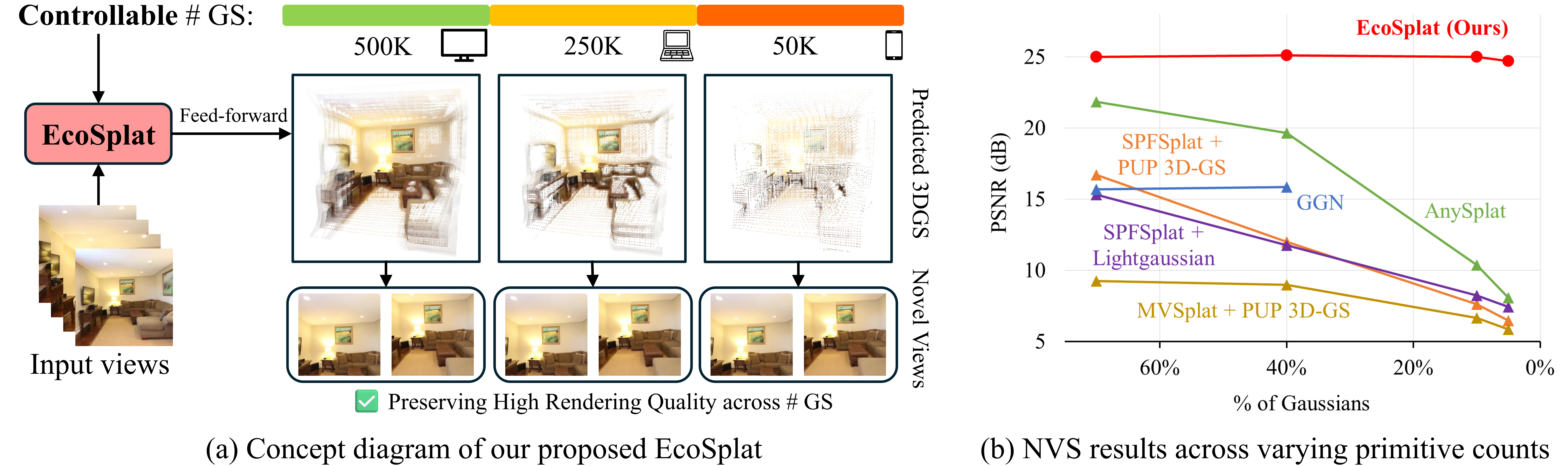}
\vspace{-0.3cm}
\captionof{figure}{{(a) EcoSplat is a feed-forward 3D Gaussian Splatting framework that enables explicit control over the number of output primitives. (b) It consistently outperforms state-of-the-art methods across a wide range of target primitive counts.}}
\label{fig:figure_page1}
}]

\begin{abstract}
Feed-forward 3D Gaussian Splatting (3DGS) enables efficient one-pass scene reconstruction, providing 3D representations for novel view synthesis without per-scene optimization. However, existing methods typically predict pixel-aligned primitives per-view, producing an excessive number of primitives in dense-view settings and offering no explicit control over the number of predicted Gaussians. To address this, we propose EcoSplat, the first efficiency-controllable feed-forward 3DGS framework that adaptively predicts the 3D representation for any given target primitive count at inference time. EcoSplat adopts a two-stage optimization process. The first stage is Pixel-aligned Gaussian Training (PGT) where our model learns initial primitive prediction. The second stage is Importance-aware Gaussian Finetuning (IGF) stage where our model learns rank primitives and adaptively adjust their parameters based on the target primitive count. Extensive experiments across multiple dense-view settings show that EcoSplat is robust and outperforms state-of-the-art methods under strict primitive-count constraints, making it well-suited for flexible downstream rendering tasks.

\end{abstract}

{
  \renewcommand{\thefootnote}%
    {\fnsymbol{footnote}}
  \footnotetext[1]{Co-first authors (equal contribution).}
  \footnotetext[2]{Co-corresponding authors.}
}

\section{Introduction}
\label{sec:intro}
Novel View Synthesis (NVS) is a fundamental task in 3D vision, enabling applications in virtual reality (VR), augmented reality (AR), and cinematic production. Neural implicit representations~\cite{mildenhall2020nerf, barron2021mip, barron2022mip, fridovich2022plenoxels, yu2021plenoctrees, muller2022instant, cao2023hexplane} have shown impressive progress, yet their training and rendering remain computationally expensive. Recently, 3D Gaussian Splatting (3DGS)~\cite{kerbl20233d} has emerged as a state-of-the-art method to achieve both improved rendering quality and real-time rendering speed. Although 3DGS enables real-time rendering and fast reconstructions, it still requires substantial training time, often several minutes or even hours per scene for reconstruction, since it relies on per-scene optimization. To address this limitation, several feed-forward-based approaches~\cite{charatan2024pixelsplat, szymanowicz2024splatter, chen2024mvsplat, xu2025depthsplat, kang2025selfsplat, ye2024no, huang2025no, zhang2024gs} have been proposed to directly predict the properties of 3D Gaussians (primitives) from multi-view images, leveraging large-scale datasets. In general, these approaches unproject the 2D pixels of each input image into a corresponding 3D primitive (i.e., \textit{pixel-aligned} Gaussians) and aggregate the resulting primitives from all input views to represent the corresponding scene.

After the feed-forward reconstruction stage, the predicted 3D Gaussian primitives can be stored and reused for real-time rendering on end-user devices such as mobile phones or AR/VR headsets. In practical applications, reconstruction is typically performed once on the server-side devices, while the generated 3DGS must support multiple rendering requests across diverse end-user devices. However, existing feed-forward methods typically generate a large number of pixel-aligned primitives, especially for dense-view reconstructions, which can be impractical for resource-constrained devices. To mitigate this, several approaches~\cite{jiang2025anysplat, zhang2024gaussian, ziwen2025long} have attempted to reduce redundant Gaussians across multiple views through voxelization~\cite{jiang2025anysplat}, graph-based aggregation~\cite{zhang2024gaussian}, or opacity-based pruning~\cite{ziwen2025long}. Nonetheless, these methods \textit{lack explicit control} over the target numbers of primitives \textit{because they were not specifically designed for fine-level control over this number}, which is a crucial factor for flexible deployment in real-time and streamable applications. Their threshold-based pruning strategies (e.g., voxel size or opacity-based pruning) often yield scene-dependent and unstable primitive counts, resulting in a poor quality-efficiency trade-off. In practice, \textit{explicit} control over the number of 3D Gaussian primitives is essential: without a mechanism to enforce a prescribed count, the final number of primitives varies across scenes and view counts, leading to unpredictable latency, memory, and bandwidth requirements. \textit{Importantly}, it should be noted that our work is distinguished from the standardization activities of MPEG~\cite{ISO2025} and other academic works~\cite{song2025tinysplat, chen2024fast} where 3D Gaussians are subject to \textit{compression} via codecs while our approach is related to effective and controllable \textit{compaction} of feed-forward 3DGS representations in terms of the number of the primitives.

To overcome these challenges, we propose EcoSplat, an \underline{E}fficiency-\underline{co}ntrollable feed-forward 3D Gaussian \underline{Splat}ting framework for dense-view reconstruction. As shown in Fig.~\ref{fig:figure_page1}, EcoSplat maximizes rendering quality while explicitly controlling the number of 3D Gaussians to match a target number, all in a feed-forward manner. To achieve this, we introduce a two-stage training process consisting of Pixel-aligned Gaussian Training (PGT) and Importance-aware Gaussian Finetuning (IGF). In the first stage (PGT), we train our model to reliably predict pixel-aligned 3D Gaussians. In the second stage (IGF), we further train our model to adaptively predict the 3D Gaussian parameters conditioned on the target number of primitives. Specifically, in the IGF stage, we introduce an importance-aware opacity loss $\mathcal{L}_\text{io}$ to encourage the suppression of less informative primitives. Moreover, for improved training stability, we adopt a Progressive Learning on Gaussian Compaction (PLGC) strategy that gradually expands the sampling range of the target number of primitives during training. As a result of this training process, our EcoSplat can adaptively select the most important Gaussians for any target number of primitives during inference (Fig.~\ref{fig:figure_page1}-(b)). Our main contributions are summarized as follows:
\begin{itemize}
  \setlength\itemsep{0.1cm}
  \item We introduce EcoSplat, the \textit{first} feed-forward 3DGS framework that reconstructs a high-quality and robust set of Gaussian primitives while satisfying \textit{explicit and controllable constraints on the target number of primitives}.
  \item We design a two-stage training process, including {Pixel-aligned Gaussian Training (PGT)} followed by {Importance-aware Gaussian Finetuning (IGF)}, that enables the model to learn an \textit{importance-aware primitive ranking} and achieve \textit{controllable efficiency}.
  \item We demonstrate state-of-the-art efficiency-controllable NVS performance on dense-view benchmarks such as RealEstate10K~\cite{zhou2018stereo} and ACID~\cite{infinite_nature_2020}. By achieving an optimal trade-off between primitive count and rendering quality, EcoSplat significantly outperforms previous feed-forward methods that lack fine-grained control over the number of primitives.
\end{itemize}

\section{Related Work}
\label{sec:related_work}

\noindent \textbf{Optimization-based NVS.}
Since the introduction of Neural Radiance Fields (NeRF)~\cite{mildenhall2020nerf}, the research on novel view synthesis (NVS) has made remarkable progress, with numerous extensions~\cite{yu2021plenoctrees, fridovich2022plenoxels, fastnerf, efficient_nerf, muller2022instant, tensorf, cao2023hexplane, fridovich2023k, bui2025moblurf}. 3D Gaussian Splatting (3DGS)~\cite{kerbl20233d} has emerged as an explicit scene representation approach that models a scene using millions of anisotropic Gaussian primitives and employs differentiable rasterization to achieve photorealistic novel views with real-time inference. Despite these advantages, 3DGS relies on per-scene optimization of Gaussians and cannot directly render novel views from input images without retraining. Very recently, several post-optimization techniques have been proposed to control the number of Gaussian primitives~\cite{HansonTuPUP3DGS, fan2023lightgaussian}. 
PUP 3D-GS~\cite{HansonTuPUP3DGS} computes a spatial sensitivity score to prune pretrained 3DGS models.
LightGaussian~\cite{fan2023lightgaussian} utilizes significance-based pruning, SH coefficient distillation, and vector quantization to reduce the 3DGS model size.
However, similar to 3DGS~\cite{kerbl20233d}, these methods cannot operate in a feed-forward manner.

\begin{figure*}
\centering
\includegraphics[width=\linewidth,keepaspectratio]{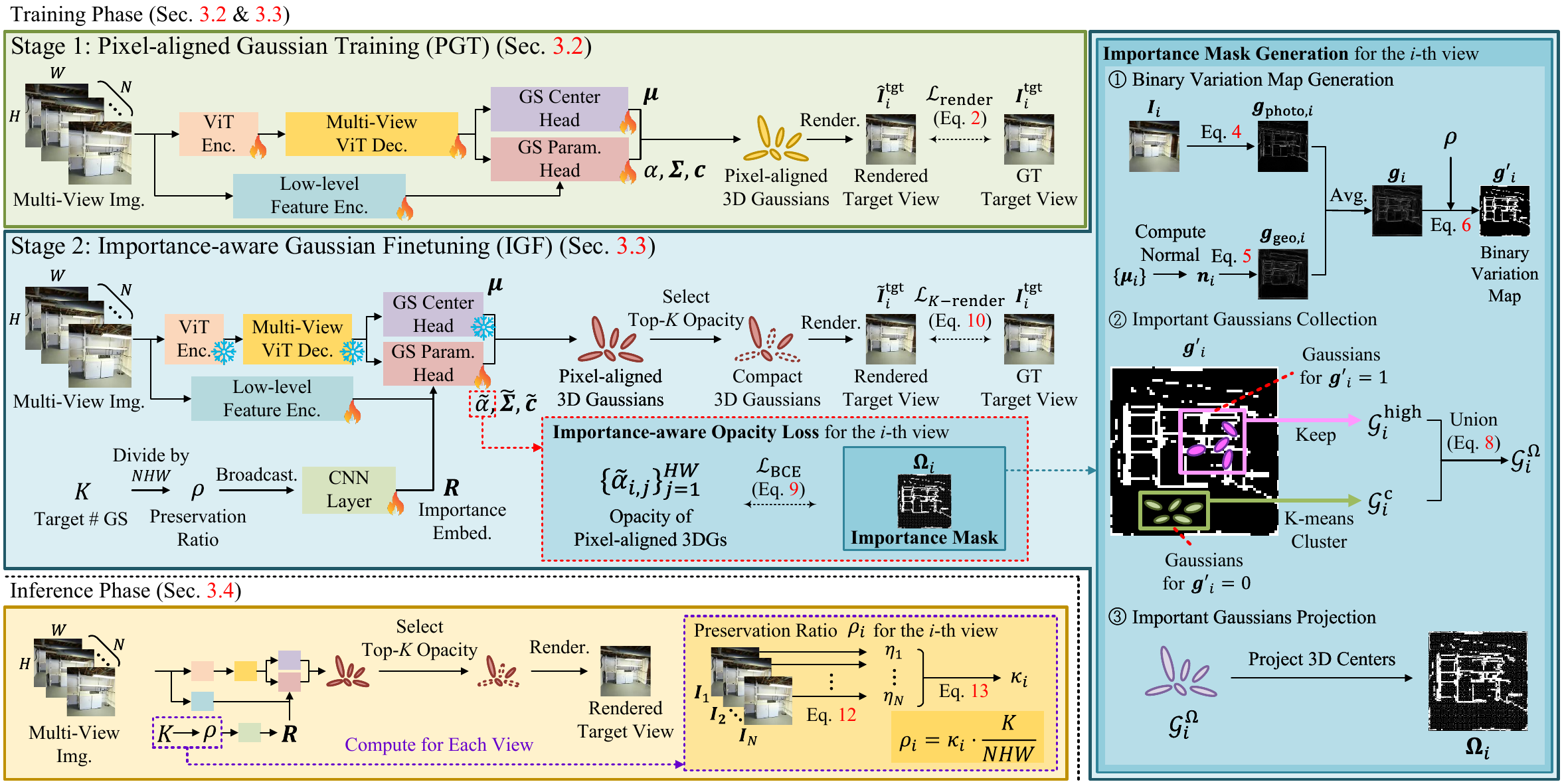}
\vspace{-0.6cm}
\caption{\textbf{Overview of EcoSplat.} EcoSplat is trained in two stages: Pixel-aligned Gaussian Training (PGT) (Sec.~\ref{sec:architecture}) and Importance-aware Gaussian Finetuning (IGF) (Sec.~\ref{sec:IGF}).
During IGF, the combination of the importance-aware opacity loss $\mathcal{L}_\text{io}$ and the Progressive Learning on Gaussian Compaction (PLGC) encourages EcoSplat to suppress the opacities of less important Gaussians. At inference, it adaptively satisfies an arbitrary user-specified primitive count and produces the optimal Gaussians in a feed-forward manner (Sec.~\ref{sec:inference}).}
\label{fig:overall_architecture}
\end{figure*}

\noindent \textbf{Feed-forward NVS.} 
To enable generalizable NVS, early methods learn generalizable neural radiance fields~\cite{pixelnerf, mvsnerf, nerf_attention} or image-based rendering models~\cite{ibrnet, neuray, gpnr}, leveraging data-driven priors via transformer-based architectures. More recently, due to the superior rendering quality and efficiency of 3DGS, several feed-forward approaches~\cite{charatan2024pixelsplat, szymanowicz2024splatter, chen2024mvsplat, xu2025depthsplat, kang2025selfsplat, ye2024no, huang2025no, zhang2024gs} have been proposed to directly generate 3D Gaussians from input images. These methods predict the parameters of a single 3D Gaussian per pixel in each input image to represent the target scene. PixelSplat~\cite{charatan2024pixelsplat} and Splatter Image~\cite{szymanowicz2024splatter} are pioneering works that predict 3D Gaussians directly from image features. MVSplat~\cite{chen2024mvsplat} constructs a cost volume via plane sweeping to capture cross-view feature correspondences, while DepthSplat~\cite{xu2025depthsplat} leverages pre-trained monocular depth features~\cite{yang2024depth} to achieve more robust 3D Gaussian reconstructions. Several methods~\cite{hong2024unifying, kang2025selfsplat, li2024ggrt, ye2024no, huang2025no} have been studied for pose-free feed-forward NVS. However, all of the aforementioned feed-forward methods are constrained by their \textit{pixel-aligned Gaussian} representations, which inevitably introduce redundant Gaussians, limiting their efficiency and scalability for dense multi-view scenarios.

\noindent \textbf{Efficient Feed-forward NVS.} 
To address the inefficiency of the above \textit{pixel-aligned} feed-forward NVS methods, recent studies~\cite{jiang2025anysplat, zhang2024gaussian, ziwen2025long, liu2025worldmirror} have proposed strategies for Gaussian aggregation and pruning. AnySplat~\cite{jiang2025anysplat} and WorldMirror~\cite{liu2025worldmirror}, built upon the VGGT~\cite{wang2025vggt} architecture, cluster Gaussians into fixed-size voxels via differentiable voxelization, but the voxel size is a sensitive hyperparameter that strongly affects rendering quality. GGN~\cite{zhang2024gaussian} models the inter-view relationships among Gaussian groups using a graph-based approach and prunes redundant Gaussians via pooling. However, this graph-based approach has been shown to lack sufficient robustness, which hinders its generalization across diverse scene geometries and dense multi-view inputs.
Long-LRM~\cite{ziwen2025long} utilizes a Mamba2~\cite{dao2024transformers}-based architecture to compress long sequences of Gaussians through token merging and opacity-based pruning. Despite of their advancements in efficient rendering, the aforementioned approaches provide neither explicit control over the desired number of primitives nor a guaranteed, optimal trade-off between rendering fidelity and primitive count.

\section{Proposed Method: EcoSplat}
\label{sec:proposed_method}

\subsection{Overview of EcoSplat}
Given $N$ multi-view images $\{\bm{I}_i\}_{i=1}^{N}$ for a scene and a target number of Gaussians $K$, we aim to predict, in a feed-forward manner, a set $\mathcal{G}=\{\bm{G}_k\}_{k=1}^{K}$ of 3D Gaussian primitives~\cite{kerbl20233d} that maximizes the rendering quality.
Our proposed {EcoSplat} controls the per-view number of predicted 3D Gaussian primitives by ranking and selecting the most informative primitives in each view, ensuring that the set of selected primitives meets the target primitive number $K$.

To accomplish this, we adopt a two-stage optimization process: (i) The first stage is Pixel-aligned Gaussian Training (PGT) (Sec.~\ref{sec:architecture}) where we train our model to predict pixel-aligned Gaussians, similar to prior works~\cite{charatan2024pixelsplat, chen2024mvsplat}; (ii) The second stage is Importance-aware Gaussian Finetuning (IGF) (Sec.~\ref{sec:IGF}) where our model is further trained to predict the Gaussian parameters (opacities, 3D covariances, and colors) adaptively based on the target primitive number $K$, enabling effective ranking and selection of important Gaussians.
Specifically, we inject $K$ as a conditioning signal into our model. Adaptive to $K$, EcoSplat suppresses the opacities of less important Gaussians in the pixel-aligned set, allowing the most informative ones to be retained. This process ensures the final output can be obtained by directly preserving the Gaussian primitives with the $K$-highest opacity scores, which are called `top-$K$ Gaussians'. For inference (Sec.~\ref{sec:inference}), we describe our procedure, which satisfies the target primitive number $K$ by allocating per-view Gaussian counts based on the importance of each view to maximize the rendering quality.

\subsection{Pixel-aligned Gaussian Training (PGT)}
\label{sec:architecture}
In the PGT stage, we train a pixel-aligned feed-forward 3DGS model whose architecture is built on the recent methods~\cite{ye2024no,huang2025no}.
Each view image $\bm{I}_i$ is tokenized into $P$ tokens. These tokens are then processed independently by a shared ViT~\cite{dosovitskiy2020image} encoder to produce encoded tokens. The encoded tokens from all views are then fed jointly to a ViT decoder composed of $m$ decoder blocks. This decoder applies cross-attention to aggregate multi-view information into a unified scene representation.
The output of the decoder for the $i$-th view consists of multiple decoded tokens $\{\bm{Z}_i^{(\ell)}\}_{\ell=1}^{m}$, where each
$\bm{Z}_i^{(\ell)} \in \mathbb{R}^{P \times C_\ell}$, with $C_\ell$ denoting the per-token channel dimension, is extracted from the $\ell$-th decoder block. These decoded tokens are fed into a Gaussian center head $F_\mu$ and a Gaussian parameter head $F_\nu$, which predict the 3D centers and all remaining parameters of the pixel-aligned Gaussians for the $i$-th view, respectively, as:
\begin{equation}
\begin{aligned}
        &\{\bm{\mu}_{i,j}\}_{j=1}^{H W} = F_\mu(\{\bm{Z}_i^{(\ell)}\}_{\ell=1}^{m}), \\
        &\{[\alpha_{i,j} ;\bm{\Sigma}_{i,j} ;\bm{c}_{i,j}]\}_{j=1}^{HW}  =F_\nu(\{\bm{Z}_i^{(\ell)}\}_{\ell=1}^{m},\psi(\bm{I}_i)),
    \label{eq:stage1_GS_params}
\end{aligned}
\end{equation}
where $\bm{\mu}_{i,j}$, $\alpha_{i,j}$, $\bm{\Sigma}_{i,j}$, and $\bm{c}_{i,j}$ represent the 3D center, opacity, 3D covariance, and color, respectively, of the 3D Gaussian $\bm{G}_{i,j}$ corresponding to the $j$-th pixel in the $i$-th view. $\psi(\bm{I}_i)$ denotes a feature map extracted from the input image $\bm{I}_i$ via a shallow CNN layer $\psi$ that acts as a low-level feature encoder. Using the predicted pixel-aligned Gaussians, derived from all $N$ input images, we render a target novel view image $\hat{\bm{I}}^\text{tgt}$ via rasterization with its corresponding camera. To train our model, we compute the rendering loss as the sum of the MSE and LPIPS~\cite{zhang2018perceptual} losses, averaged across all $N^\text{tgt}$ target novel views as:
\begin{equation}
\scalebox{0.9}{$
\mathcal{L}_\text{render} = \frac{1}{N^\text{tgt}} \sum\limits_{p=1}^{N^\text{tgt}} \mathcal{L}_\text{MSE}(\bm{I}^\text{tgt}_p, \hat{\bm{I}}^\text{tgt}_p) + 0.05 \cdot\mathcal{L}_\text{LPIPS}(\bm{I}^\text{tgt}_p, \hat{\bm{I}}^\text{tgt}_p).
$}
\label{eq:stage1_loss}
\end{equation}

\subsection{Importance-aware Gaussian Finetuning (IGF)}
\label{sec:IGF}
Our EcoSplat mitigates the redundancy of pixel-aligned prediction, where the total number of generated primitives grows linearly with both the number of input views and the image resolution, leading to inefficient rendering. In the IGF stage, we aim to make our model aware of the target primitive number $K$, enabling it to adaptively predict the parameters of the pixel-aligned 3D Gaussians. For this, we finetune the Gaussian parameter head $F_\nu$ to adaptively suppress the opacities $\{\alpha_{i,j}\}$ of less important primitives, while simultaneously optimizing $\{\bm{\Sigma}_{i,j}\}$ and $\{\bm{c}_{i,j}\}$ to maximize rendering quality. We inject a learnable importance embedding $\bm{R}_i \in \mathbb{R}^{H \times W \times C}$, which is derived from the target primitive number $K$, into the intermediate feature of $F_\nu$. Then, $F_\nu$ predicts the adjusted per-view pixel-aligned 3D Gaussian parameters as:
\begin{equation}
\scalebox{0.95}{$
\{[\tilde{\alpha}_{i,j} ; \tilde{\bm{\Sigma}}_{i,j} ; \tilde{\bm{c}}_{i,j}]\}_{j=1}^{HW} 
= F_\nu(\{\bm{Z}_i^{(\ell)}\}_{\ell=1}^{m}, \psi(\bm{I}_i), \bm{R}_i ),
$}
\label{eq:stage2_GS_params}
\end{equation}
where $\tilde{(\cdot)}$ denotes the adjusted Gaussian parameters. To predict $\bm{R}_i$, we compute the preservation ratio $\rho_i$ from $K$ as $\rho_i = \frac{K}{NHW}$ and broadcast $\rho_i$ to an $H\times W$ tensor, and feed it into a shallow CNN layer. Note that $\rho_i$ is set to a uniform value for all $N$ views during training, while it is computed individually for each view during inference to maximize the rendering quality (Sec.~\ref{sec:inference}). In the IGF stage, the ViT encoders/decoders and the Gaussian center head $F_\mu$ are kept frozen as shown in Fig.~\ref{fig:overall_architecture}.  

\noindent \textbf{Importance Mask Generation.}
To adaptively finetune $F_\nu$ with respect to $K$, we aim to generate a pseudo ground truth (pseudo-GT) 2D importance mask $\Omega_i$ for each view. This $\Omega_i$ can be inferred from the set $\mathcal{G}_i^\Omega$ of important 3D Gaussians that should be retained. The importance mask generation process is divided into three steps, as follows:

\noindent \textit{1) Binary Variation Map Generation}:
First, we compute a variation map $\bm{g}_i \in \mathbb{R}^{H \times W}$ for $\bm{I}_i$ that captures both photometric and geometric complexities at each pixel. The photometric variation map $\bm{g}_{\text{photo},i} \in \mathbb{R}^{H \times W}$ is computed as the gradient magnitude of $\bm{I}_i$:
\begin{equation}
    \bm{g}_{\text{photo},i} = \sqrt{\|\nabla_x \bm{I}_i\|_2^2 + \|\nabla_y \bm{I}_i\|_2^2},
\label{eq:photometric_variation}
\end{equation}
where $\|\cdot\|_2$ denotes the L2-norm. To compute the geometric variation map $\bm{g}_{\text{geo},i} \in \mathbb{R}^{H \times W}$, we obtain a depth map from the predicted means $\{\bm{\mu}_{i,j}\}_{j=1}^{H W}$ of the pixel-aligned 3D Gaussians and  compute its corresponding normal map $\bm{n}_i$. Then, $\bm{g}_{\text{geo},i}$ is computed from $\bm{n}_i$ as:
\begin{equation}
\bm{g}_{\text{geo},i} = \sqrt{\|\nabla_x \bm{n}_i\|_2^2 + \|\nabla_y \bm{n}_i\|_2^2}.
\end{equation}
The final variation map $\bm{g}_i$ is the average of the two variation maps, $\bm{g}_i = (\bm{g}_{\text{photo},i} + \bm{g}_{\text{geo},i})/2$. We then binarize $\bm{g}_i$, where $\bm{g}_{i,j}$ is the variation value at the $j$-th pixel location, to obtain a binary variation map $\bm{g}'_i$ as:
\begin{equation}
\scalebox{0.9}{$
\bm{g}'_{i,j} =
\begin{cases}
1, & \text{if} \; \bm{g}_{i,j} > \epsilon_i, \\
0, & \text{otherwise},
\end{cases}
\quad \text{where} \quad
\epsilon_i = Q_{\rho_i}(\{\bm{g}_{i,j}\}_{j=1}^{HW}),
$}
\end{equation}
where $Q_{\rho_i}(\cdot)$ denotes the $\rho_i$-th quantile. Here, $\bm{g}'_{i,j}=1$ indicates high-variation pixels and $0$ othewise.

\noindent \textit{2) Important Gaussians Collection}: Using the obtained binary variation map $\bm{g}'_i$, we select the set $\mathcal{G}_i^\Omega$ of important 3D Gaussians. We partition the predicted pixel-aligned 3D Gaussians from $\bm{I}_i$ using $\bm{g}'_{i}$ into two sets of 3D Gaussians:
\begin{equation}
\scalebox{0.90}{$
\mathcal{G}^{\text{high}}_i = \{\bm{G}_{i,j} \mid \bm{g}'_{i,j}=1\}, \quad
\mathcal{G}^{\text{low}}_i = \{\bm{G}_{i,j} \mid \bm{g}'_{i,j}=0\}.
$}
\end{equation}
Intuitively, high-variation pixels should be represented by individual primitives to preserve fine reconstruction details. Therefore, we fully include $\mathcal{G}^{\text{high}}_i$ within the set of important 3D Gaussians $\mathcal{G}^{\Omega}_i$, as $\mathcal{G}^{\Omega}_i \supseteq \mathcal{G}^{\text{high}}_i$. Next, we reduce the redundancy in the low-variation set $\mathcal{G}^{\text{low}}_i$ by merging its Gaussians using single-step K-means clustering. We first partition the input image $\bm{I}_i$ into non-overlapping $4\times4$ patches. Within the low-variation set $\mathcal{G}^{\text{low}}_i$, we designate the Gaussian associated with the first (top-left) pixel of each patch as a \emph{key} Gaussian. The centers of these key Gaussians serve as cluster centroids. We then run single-step K-means clustering over $\mathcal{G}^{\text{low}}_i$ to compute the final compact set $\mathcal{G}^{\text{c}}_i$. This procedure reduces redundancy while preserving local structure, yielding a smaller, faithful set of Gaussians for subsequent processing. The final set $\mathcal{G}_i^{\Omega}$ is computed as:
\begin{equation}
\mathcal{G}_i^{\Omega} \triangleq \mathcal{G}_i^{\text{high}} \cup \mathcal{G}_i^{\text{c}}.
\end{equation}


\noindent \textit{3) Important Gaussians Projection}:
The per-view importance mask $\Omega_i$ is generated by projecting the 3D centers of all Gaussians in $\mathcal{G}^{\Omega}_i$ onto the $i$-th image plane. This results in a binary mask where pixels intersected by any projected center are set to 1, and all other pixels are set to 0.

\noindent \textbf{Loss Function.} The model needs to learn suppressing the opacity of less important Gaussians while retaining those of more important Gaussians. For this, we introduce an importance-aware opacity loss $\mathcal{L}_{\text{io}}$ that encourages $\{\tilde{\alpha}_{i,j}\}_{j=1}^{HW}$ to match the importance mask $\Omega_i$ using a binary cross-entropy (BCE) loss as:
\begin{equation}
    \mathcal{L}_{\text{io}}
    = \lambda_{\text{io}} \cdot \frac{1}{NHW}
      \sum_{i=1}^{N} \sum_{j=1}^{HW}
      \mathcal{L}_{\text{BCE}}\!\big(\Omega_{i,j},\, \tilde{\alpha}_{i,j}\big).
\end{equation}
Finally, unlike the PGT stage, we only use the top-$K$ Gaussians to render a target novel view image $\tilde{\bm{I}}^\text{tgt}$ via rasterization. The rendering loss is then computed as (similar to Eq.~\ref{eq:stage1_loss}):
\begin{equation}
\scalebox{0.84}{$
    \mathcal{L}_{K\text{-render}} = \frac{1}{N^\text{tgt}} \sum\limits_{p=1}^{N^\text{tgt}} \mathcal{L}_\text{MSE}(\bm{I}^\text{tgt}_p, \tilde{\bm{I}}^\text{tgt}_p)  + 0.05 \cdot\mathcal{L}_\text{LPIPS}(\bm{I}^\text{tgt}_p, \tilde{\bm{I}}^\text{tgt}_p).
$}
\label{eq:stage2_loss}
\end{equation}
The total loss is computed as $\mathcal{L} = \mathcal{L}_\text{io} + \mathcal{L}_{K\text{-render}}$. Based on this total loss, our model is trained to predict an optimal set of primitives that maximizes rendering quality while adhering to the target primitive number $K$.

\noindent \textbf{Progressive Learning on Gaussian Compaction (PLGC).} It is important to stably train our model such that it can be robust to meet a wide range of target primitive numbers $K$. A naive approach for this is to randomly select $K$ values over the entire training process. However, we found that this can lead to unstable training. To mitigate this, we propose a progressive learning strategy that expands the sampling range of $K$ depending on the current training iteration. Specifically, we define the sampling interval as $[K_{\text{min}}, K_{\text{max}}]$, where the upper bound is fixed to $K_{\text{max}} = 0.95 \cdot NHW$. $K_{\text{min}}$ is gradually annealed from $0.85 \cdot NHW$ down to $0.05 \cdot NHW$ according to:
\begin{equation}
    K_{\text{min}} = \max\big(0.85 - \lambda_\text{decay} \lfloor t / S \rfloor, \, 0.05\big) \cdot NHW,
\end{equation}
where $t$ denotes the current iteration, $S$ is the decay interval, and $\lambda_\text{decay}$ is the decay rate. At each iteration, $K$ is randomly sampled from the interval $[K_{\text{min}}, K_{\text{max}}]$. This sampling strategy allows the model to gradually adapt to more aggressive pruning, ensuring both stability and effectiveness across a wide range of target primitive numbers.

\subsection{Inference}
\label{sec:inference}
During inference, to predict the target $K$ Gaussians, we allocate different numbers of primitives to each view, as some views contain more informative Gaussians than others. As the images with large portions of complex details tend to contain more informative primitives for scene details, more primitives are necessitated for such images. In this regard, we incorporate a view importance factor $\kappa_i$ as the proportion of high-frequency components to indicate the amount of scene details in the $i$-th view.
Thus, the view-specific preservation ratio $\rho_i$ can be adaptively determined based on the view importance factor $\kappa_i$ as $\rho_i = \kappa_i \rho$, where $\rho = \frac{K}{NHW}$.
This adaptive allocation allows our EcoSplat to focus resources on the views that are more critical for reconstruction quality, while reducing redundancy in less informative views.
Following MoBGS~\cite{bui2025mobgsmotiondeblurringdynamic}, we quantify the high-frequency components of each image $\bm{I}_i$ by defining a high-frequency score $\eta_i$ as the ratio of spectral magnitude in the high-frequency region to the total spectral magnitude of the $i$-th image.
We apply a 2D Discrete Fourier Transform (DFT) to $\bm{I}_i$ and shift the resulting spectrum to center the low-frequency components, yielding the shifted DFT $\tilde{\mathcal{F}}(\bm{I}_i)$.
The high-frequency score $\eta_i$ is then computed as:
\begin{equation}
    \eta_i = 1 - \frac{\sum_{\bm{\xi} \in \Lambda} E_i(\bm{\xi})}{\sum_{\bm{\xi}} E_i(\bm{\xi})},
    \quad \text{where} \;\; E_i = |\tilde{\mathcal{F}}(\bm{I}_i)|,
\label{eq:high_frequency_score}
\end{equation}
where $\bm{\xi} = (w_x,w_y)$ denotes a 2D frequency index vector with horizontal and vertical frequencies of $w_x$ and $w_y$, and $\Lambda$ is the centered square region of side length $s$ that captures the low-frequency portion of $\tilde{\mathcal{F}}(\bm{I}_i)$.
Then, we compute the view-wise importance factor as:
\begin{equation}
    \kappa_i = N\cdot\Psi_i, \quad \text{where}  \;\; \Psi_i = \frac{e^{\eta_i / T}}{\sum_{q=1}^{N} e^{\eta_q / T}}.
\label{eq:view_wise_impotance_factor}
\end{equation}
With temperature $T$, this softmax formulation amplifies differences among the high-frequency scores $\eta_i$ across views.
We then set $\rho_i = \kappa_i\,\rho$, thereby ensuring that the average preservation ratio over all input images is equal to $\rho$.
\section{Experiments}
\label{sec:experiments}
\begin{table*}
\begin{center}
\setlength\tabcolsep{6pt} 
\renewcommand{\arraystretch}{1.1}
\scalebox{0.8}{
\begin{tabular}{ l | c c c | c c c | c c c | c c c}
\toprule
\multicolumn{1}{c|}{\multirow{2}{*}{Methods}} & \multicolumn{3}{c|}{5\% Gaussians}  & \multicolumn{3}{c|}{10\% Gaussians}  & \multicolumn{3}{c|}{40\% Gaussians} & \multicolumn{3}{c}{70\% Gaussians} \\
\cline{2-13}
& PSNR$\uparrow$ & SSIM$\uparrow$ & LPIPS$\downarrow$ & PSNR$\uparrow$ & SSIM$\uparrow$ & LPIPS$\downarrow$ & PSNR$\uparrow$ & SSIM$\uparrow$ & LPIPS$\downarrow$ & PSNR$\uparrow$ & SSIM$\uparrow$ & LPIPS$\downarrow$ \\
\midrule
GGN~\cite{zhang2024gaussian} & N/A & N/A & N/A & N/A & N/A & N/A & 15.86 & 0.580 & 0.408 & 15.71 & 0.577 & 0.408 \\
AnySplat~\cite{jiang2025anysplat} & 8.08 & \second{0.229} & 0.593 & \second{10.37} & \second{0.363} & \second{0.513} & \second{19.66} & 0.677 & \second{0.248} & 21.85 & 0.723 & \second{0.177} \\
WorldMirror~\cite{liu2025worldmirror} & \second{8.09} & 0.220 & 0.632 & 10.20 & 0.349 & 0.552 & \second{19.66} & 0.692 & 0.277 & \second{22.11} & \second{0.744 }& 0.196 \\
MVSplat~\cite{chen2024mvsplat} +~\cite{fan2023lightgaussian} & 5.82 & 0.053 & 0.717 & 6.41 & 0.085 & 0.696 & 8.71 & 0.224 & 0.598 & 9.26 & 0.245 & 0.581 \\
MVSplat~\cite{chen2024mvsplat} +~\cite{HansonTuPUP3DGS} & 5.86 & 0.051 & 0.699 & 6.66 & 0.096 & 0.678 & 9.00 & 0.232 & 0.590 & 9.26 & 0.245 & 0.580 \\
SPFSplat~\cite{huang2025no} +~\cite{fan2023lightgaussian} & 7.44 & 0.231 & 0.624 & 8.23 & 0.280 & 0.584 & 11.77 & 0.514 & 0.410 & 15.32 & 0.671 & 0.280 \\
SPFSplat~\cite{huang2025no} +~\cite{HansonTuPUP3DGS}& 6.48 & 0.167 & \second{0.589} & 7.61 & 0.256 & 0.536 & 12.01 & 0.522 & 0.378 & 16.72 & 0.697 & 0.257\\
\midrule
EcoSplat (Ours) & \best{24.72} & \best{0.822} & \best{0.183} & \best{25.00} & \best{0.831} & \best{0.171} & \best{25.11} & \best{0.835} & \best{0.164} & \best{25.00} & \best{0.832} & \best{0.166} \\
\bottomrule
\end{tabular}
}
\vspace{-0.2cm}
\caption{\textbf{Quantitative comparison of NVS under controlled numbers of primitives on the RE10K dataset~\cite{zhou2018stereo} with 24 input views.} 
\best{Red} and \second{Blue} indicate the best and second-best performances, respectively. `N/A' denotes that GGN~\cite{zhang2024gaussian} cannot achieve configurations with primitive counts below 15\% of the total number of pixel-aligned Gaussians due to its limited control over the primitive count.}
\label{table:primitive_ratio}
\vspace{-0.4cm}
\end{center}
\end{table*}

\begin{figure*}[t]
    \centering    \includegraphics[width=0.98\linewidth,keepaspectratio]{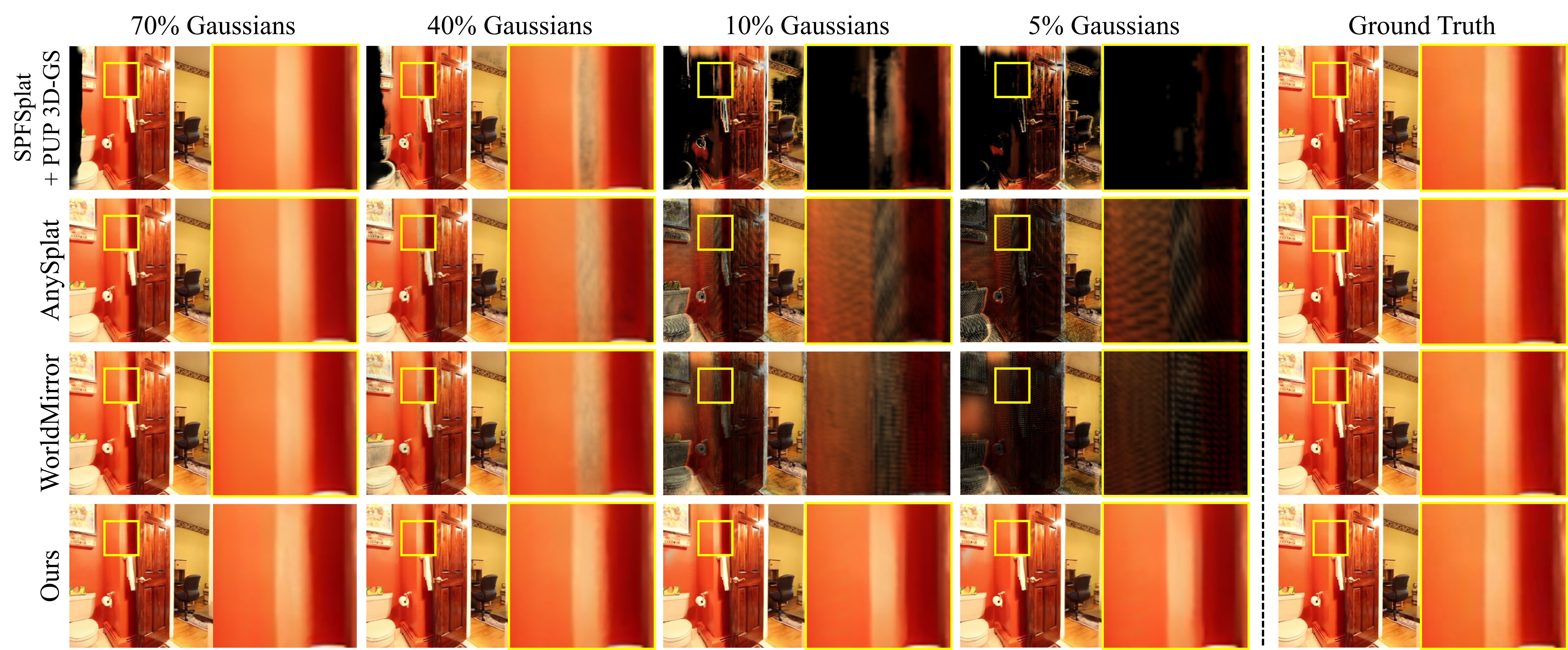}
    \caption{\textbf{Visual comparison of NVS under controlled numbers of primitives on the RE10K dataset~\cite{zhou2018stereo} with 24 input views.}}
    \label{fig:qualitative_varying_K}
\end{figure*}

\noindent \textbf{Implementation Details.}
EcoSplat is implemented in PyTorch~\cite{Khvedchenya_Eugene_2019_PyTorch_Toolbelt} using the gsplat rasterization engine~\cite{ye2025gsplat}. Our training is conducted on 4 NVIDIA A100 GPUs (40GB each). Following prior works~\cite{huang2025no, ye2024no}, the ViT encoder/decoder and $F_\mu$ are initialized with pretrained weights from MASt3R~\cite{mast3r_eccv24}. For both PGT and IGF stages, we set the number of iterations to 200K. The number of decoder blocks $m$ is set to 4. For $\mathcal{L}_\text{io}$, we set $\lambda_\text{io}$ to 0.1. For PLGC, we set $\lambda_\text{decay}$ and $S$ to 0.05 and 1000, respectively. For Eq.~\ref{eq:high_frequency_score}, we set the side length of crop region $s$ to 64. For Eq.~\ref{eq:view_wise_impotance_factor}, we set $T$ to 0.2.


\noindent \textbf{Datasets.} 
We train our EcoSplat model on the large-scale {RealEstate10K (RE10K)} dataset~\cite{zhou2018stereo}, which contains over {10 million frames} collected from approximately {10K YouTube real estate videos}. To assess our model's rendering quality, we use the RE10K test split and the ACID dataset~\cite{infinite_nature_2020} for in-domain and cross-domain generalization, respectively. The ACID dataset~\cite{infinite_nature_2020} consists of thousands of aerial drone YouTube videos depicting diverse coastal and natural scenes. Following prior works~\cite{xu2025depthsplat, chen2024mvsplat, ye2024no, huang2025no}, our EcoSplat model is trained and evaluated at an image resolution of $256\times256$.



\noindent \textbf{Evaluation Protocol.}  
We follow the evaluation protocol of NoPoSplat~\cite{ye2024no}. To evaluate model performance with a large number of input views, we first select challenging sequences from {NoPoSplat} characterized by large spatial gaps (0.05\%-0.3\% image overlap). We then densify these sparse sequences by randomly sampling intermediate views to meet our desired numbers of input views (e.g., 16, 20, 24 views). Following prior works~\cite{xu2025depthsplat,jiang2025anysplat, huang2025no, zhang2024gaussian, ye2024no}, we quantitatively evaluate NVS quality using Peak Signal-to-Noise Ratio (PSNR), Structural Similarity Index Measure (SSIM), and Learned Perceptual Image Patch Similarity (LPIPS)~\cite{zhang2018perceptual}. The final number of Gaussians is also reported as an additional metric to highlight EcoSplat's robustness when rendering for various numbers of primitives.



\begin{table*}
\begin{center}
\setlength\tabcolsep{6pt} 
\renewcommand{\arraystretch}{1.1}
\scalebox{0.8}{
\begin{tabular}{ l | c c c c | c c c c | c c c c}
\toprule
\multicolumn{1}{c|}{\multirow{2}{*}{Methods}} & \multicolumn{4}{c|}{16 Views}  & \multicolumn{4}{c|}{20 Views}  & \multicolumn{4}{c}{24 Views} \\
\cline{2-13}
& PSNR$\uparrow$ & SSIM$\uparrow$ & LPIPS$\downarrow$ & \# GS$\downarrow$ & PSNR$\uparrow$ & SSIM$\uparrow$ & LPIPS$\downarrow$ & \# GS$\downarrow$ & PSNR$\uparrow$ & SSIM$\uparrow$ & LPIPS$\downarrow$ & \# GS$\downarrow$ \\
\midrule
MVSplat~\cite{chen2024mvsplat} & 14.88 & 0.463 & 0.435 & 1048K & 14.87 & 0.462 & 0.437 & 1310K & 14.86 & 0.461 & 0.440 & 1573K \\
DepthSplat~\cite{xu2025depthsplat} & 20.05 & 0.752 & 0.258 & 1048K & 19.53 & 0.733 & 0.274 & 1310K & 19.18 & 0.718 & 0.286 & 1573K \\
NoPoSplat~\cite{ye2024no} & 13.66 & 0.450 & 0.488 & 1048K & 13.47 & 0.441 & 0.496 & 1310K & 14.38 & 0.444 & 0.489 & 1573K \\
SPFSplat~\cite{huang2025no} & \second{25.09} & \best{0.841} & \best{0.141} & 1048K & \second{24.97} & \best{0.838} & \best{0.142} & 1310K & \second{24.74} & \second{0.832} & \best{0.145} & 1573K \\
GGN~\cite{zhang2024gaussian} & 15.82 & 0.578 & 0.409 & 425K & 15.81 & 0.578 & 0.408 & \second{472K} & 15.80 & 0.578 & 0.408 & \second{512K} \\
AnySplat~\cite{jiang2025anysplat} & 21.73 & 0.720 & 0.177 & 887K & 21.88 & 0.725 & 0.173 & 1078K & 21.90 & 0.724 & 0.173 & 1259K \\
WorldMirror~\cite{liu2025worldmirror} & 21.96 & 0.737 & 0.197 & 746K & 22.07 & 0.741 & 0.195 & 891K & 22.16 & 0.745 & 0.193 & 1020K \\ 
\midrule
Ours$_{5\%}$ & 24.77 & 0.823 & 0.182 &  \best{52K} & 24.78 & 0.825 & 0.181 &  \best{65K} & 24.72 & 0.822 & 0.183 &  \best{78K} \\
Ours$_{40\%}$ & \best{25.27} & \second{0.839} & \second{0.161} &  \second{419K} & \best{25.21} & \second{0.837} & \second{0.162} &  {524K} & \best{25.11} & \best{0.835} & \second{0.164} &  {629K} \\
\bottomrule
\end{tabular}
}
\vspace{-0.1cm}
\caption{\textbf{Quantitative comparison of NVS on the RE10K dataset~\cite{zhou2018stereo} under various input-view settings.} We evaluate all models with 16, 20, and 24 input views. `Ours$_{5\%}$' and `Ours$_{40\%}$' refer to our EcoSplat model constrained to 5\% and 40\% of the total pixel-aligned Gaussians, respectively. We provide the inference FPS for all baselines in the \textit{Supplementary}.}
\label{table:re10k_multi_view}
\vspace{-0.4cm}
\end{center}
\end{table*}


\begin{table}
\begin{center}
\setlength\tabcolsep{6pt} 
\renewcommand{\arraystretch}{1.1}
\scalebox{0.8}{
\begin{tabular}{ l | c c c c}
\toprule
\multicolumn{1}{c|}{Methods} & PSNR$\uparrow$ & SSIM$\uparrow$ & LPIPS$\downarrow$ & \# GS$\downarrow$\\
\midrule
MVSplat~\cite{chen2024mvsplat} & 18.09 & 0.500 & 0.379 & 1573K \\
DepthSplat~\cite{xu2025depthsplat} & 19.78 & 0.688 & 0.289 & 1573K \\
NoPoSplat~\cite{ye2024no} & 16.87 & 0.496 & 0.413 & 1573K \\
SPFSplat~\cite{huang2025no} & \best{24.40} & \best{0.720} & \best{0.222} & 1573K \\
GGN~\cite{zhang2024gaussian} & 18.06 & 0.553 & 0.389 & \second{475K} \\
AnySplat~\cite{jiang2025anysplat} & 21.96 & 0.619 & 0.258 & 1230K \\
WorldMirror~\cite{liu2025worldmirror} & 22.34 & 0.658 & 0.273 & 1141K \\
\midrule
Ours$_{5\%}$ & 23.87 & 0.689 & 0.282 & \best{78K} \\
Ours$_{40\%}$ & \second{24.02} & \second{0.696} & \second{0.256} & 629K \\
\bottomrule
\end{tabular}
}
\caption{\textbf{Cross-dataset generalization with 24 input views.} We train all methods on the RE10K dataset~\cite{zhou2018stereo} and evaluate their zero-shot performance on the ACID dataset~\cite{infinite_nature_2020}. We provide results for 16 and 20 views in the \textit{Supplementary}.}
\label{table:acid_multi_view}
\vspace{-0.6cm}
\end{center}
\end{table}

\subsection{Comparison with State-of-the-Art Methods}
\noindent \textbf{NVS under Controlled Number of Primitives.}
Table~\ref{table:primitive_ratio} and Fig.~\ref{fig:qualitative_varying_K} present quantitative and qualitative comparisons of feed-forward 3DGS models under varying target 3D Gaussian counts, which are set to 5\%, 10\%, 40\%, and 70\% of the total pixel-aligned primitives ($NHW$). For this comparison, we use the RE10K dataset with 24 input views.
We compare our EcoSplat with representative efficient feed-forward 3DGS methods, such as GGN~\cite{zhang2024gaussian}, AnySplat~\cite{jiang2025anysplat}, and WorldMirror~\cite{liu2025worldmirror}. Since these methods lack \textit{explicit} control over the primitive counts at inference, we tune their relevant hyperparameters to \textit{indirectly} control the primitive counts, such as similarity thresholds for GGN~\cite{zhang2024gaussian} and voxel sizes for AnySplat~\cite{jiang2025anysplat} and WorldMirror~\cite{liu2025worldmirror}.
Furthermore, we compare our EcoSplat against the cascades of the SOTA pixel-aligned feed-forward 3DGS methods (SPFSplat~\cite{huang2025no} and MVSplat~\cite{chen2024mvsplat}) and the post-pruning methods~\cite{fan2023lightgaussian, HansonTuPUP3DGS}.
To preserve the feed-forward setting, we apply only the pruning step from~\cite{fan2023lightgaussian,HansonTuPUP3DGS} to the pixel-aligned Gaussians predicted by SPFSplat~\cite{huang2025no} and MVSplat~\cite{chen2024mvsplat}, with no additional optimization. As shown in Table~\ref{table:primitive_ratio}, existing methods experience substantial degradation as the number of allowed primitives decreases. Due to its limited algorithmic design, GGN~\cite{zhang2024gaussian} cannot reduce its primitive counts below a certain minimum threshold. Voxel-based approaches, such as AnySplat~\cite{jiang2025anysplat} and WorldMirror~\cite{liu2025worldmirror}, suffer from rendering artifacts at moderate pruning levels (e.g., $40\%$) and fail completely under aggressive pruning (e.g., $5\%$, $10\%$), as illustrated in Fig.~\ref{fig:qualitative_varying_K}.
Although LightGaussian~\cite{fan2023lightgaussian} and PUP 3D-GS~\cite{HansonTuPUP3DGS} perform reasonably well for zero-shot pruning on \textit{optimized}, per-scene 3DGS, they break down on feed-forward, pixel-aligned primitives. In these per-pixel representations, each Gaussian is tied to a single pixel, so pruning any primitive directly creates missing regions (Fig.~\ref{fig:qualitative_varying_K}).
In contrast, EcoSplat maintains high rendering fidelity over a wide range of primitive counts, with only marginal degradation even under aggressive reduction of primitives. These results highlight that EcoSplat shows the robustness and effectiveness of its explicit primitive-count control.

\noindent \textbf{NVS under Various Multi-view Settings.}
Table~\ref{table:re10k_multi_view} shows the NVS performance comparison of our EcoSplat against recent SOTA feed-forward 3DGS methods~\cite{xu2025depthsplat, ye2024no, huang2025no, zhang2024gaussian, ziwen2025long, jiang2025anysplat} across varying numbers of input views. For this comparison, we explicitly constrain our EcoSplat to extremely compact (5\%) and moderate (40\%) primitive ratios to demonstrate its robust performance.
As shown in Table~\ref{table:re10k_multi_view}, EcoSplat consistently achieves the best overall rendering qualities across all input-view configurations while using substantially fewer Gaussian primitives. Moreover, even with a primitive budget \textbf{$\sim$7$\times$ smaller} than the most compact baseline, GGN~\cite{zhang2024gaussian}, EcoSplat still achieves a \textbf{+9 dB} gain in PSNR and an LPIPS score that is \textbf{over 2$\times$ lower}. Compared to other efficient baselines such as AnySplat~\cite{jiang2025anysplat} and WorldMirror~\cite{liu2025worldmirror}, EcoSplat delivers \textbf{2.5-3.5 dB} higher PSNR while requiring \textbf{over 10$\times$ fewer primitives}. These results demonstrate that EcoSplat uniquely combines superior rendering quality with effective, explicit control over the primitive counts.


\noindent \textbf{Cross-Dataset Generalization NVS.}
Table~\ref{table:acid_multi_view} shows the cross-dataset generalization performance of EcoSplat under multi-view settings, compared against recent SOTA feed-forward 3DGS methods~\cite{xu2025depthsplat, ye2024no, huang2025no, zhang2024gaussian, liu2025worldmirror, jiang2025anysplat}. For this comparison, we use the RE10K dataset~\cite{zhou2018stereo} for training and the ACID dataset~\cite{infinite_nature_2020} for evaluation.
EcoSplat consistently outperforms existing efficient feed-forward models and most of the pixel-aligned 3DGS methods except SPFSplat~\cite{huang2025no} in this setting. Note that EcoSplat underperforms SPFSplat with 0.38dB lower in PSNR, but uses substantially fewer primitives (only $40\%$).

\subsection{Ablation Study}

\begin{figure}[t]
    \centering    \includegraphics[scale=0.24]{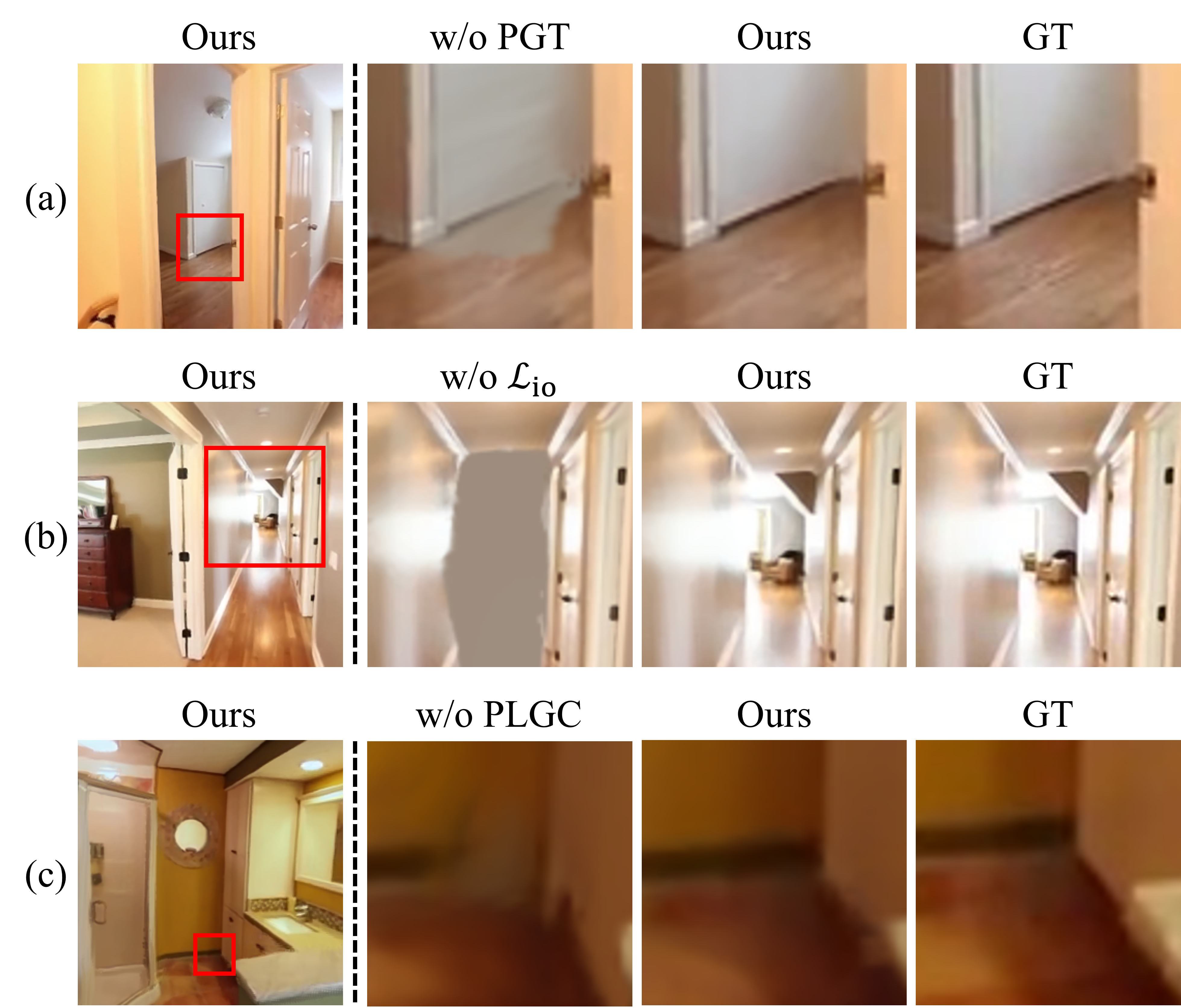}
    \vspace{-0.2cm}
    \caption{\textbf{Visual results of ablation study:} (a) the PGT stage; (b) the importance-aware opacity loss; and (c) the PLGC strategy.}
    \label{fig:ablation}
\end{figure}

\noindent \textbf{Training Strategy.}  
We validate our two-stage training strategy by ablating each stage, removing PGT and IGF to obtain the variants `w/o PGT' and `w/o IGF', respectively, as shown in Table~\ref{tab:ablation} and Fig.~\ref{fig:ablation}.
Under an extremely low primitive ratio (5\%), `w/o IGF' results in a drastic PSNR drop of over 18 dB and behaves similarly to the pixel-aligned feed-forward methods~\cite{chen2024mvsplat, huang2025no} in Table~\ref{table:primitive_ratio}.
Likewise, `w/o PGT' results in a 1.5 dB PSNR drop and visible rendering artifacts (Fig.~\ref{fig:ablation}), indicating the effectiveness of PGT.

\noindent \textbf{IGF Components.}
We vaildate the core components of the IGF stage, $\mathcal{L}_\text{io}$ and the PLGC strategy in Table~\ref{tab:ablation} and Fig.~\ref{fig:ablation}.
Without $\mathcal{L}_\text{io}$ (`w/o $\mathcal{L}_\text{io}$'), our model improperly allocates top-$K$ Gaussians relative to scene complexity, leading to under-reconstructed geometry and incomplete structural details (Fig.~\ref{fig:ablation}).
This results in significant degradation in both structural (PSNR, SSIM) and perceptual (LPIPS) metrics, especially under aggressive pruning (5\%).
Similarly, without the PLGC strategy (`w/o PLGC'), our model results in unstable optimization and fails to effectively adjust Gaussian parameters, leading to distorted outputs (Fig.~\ref{fig:ablation}).

\begin{figure}[t]
    \centering    \includegraphics[scale=0.22]{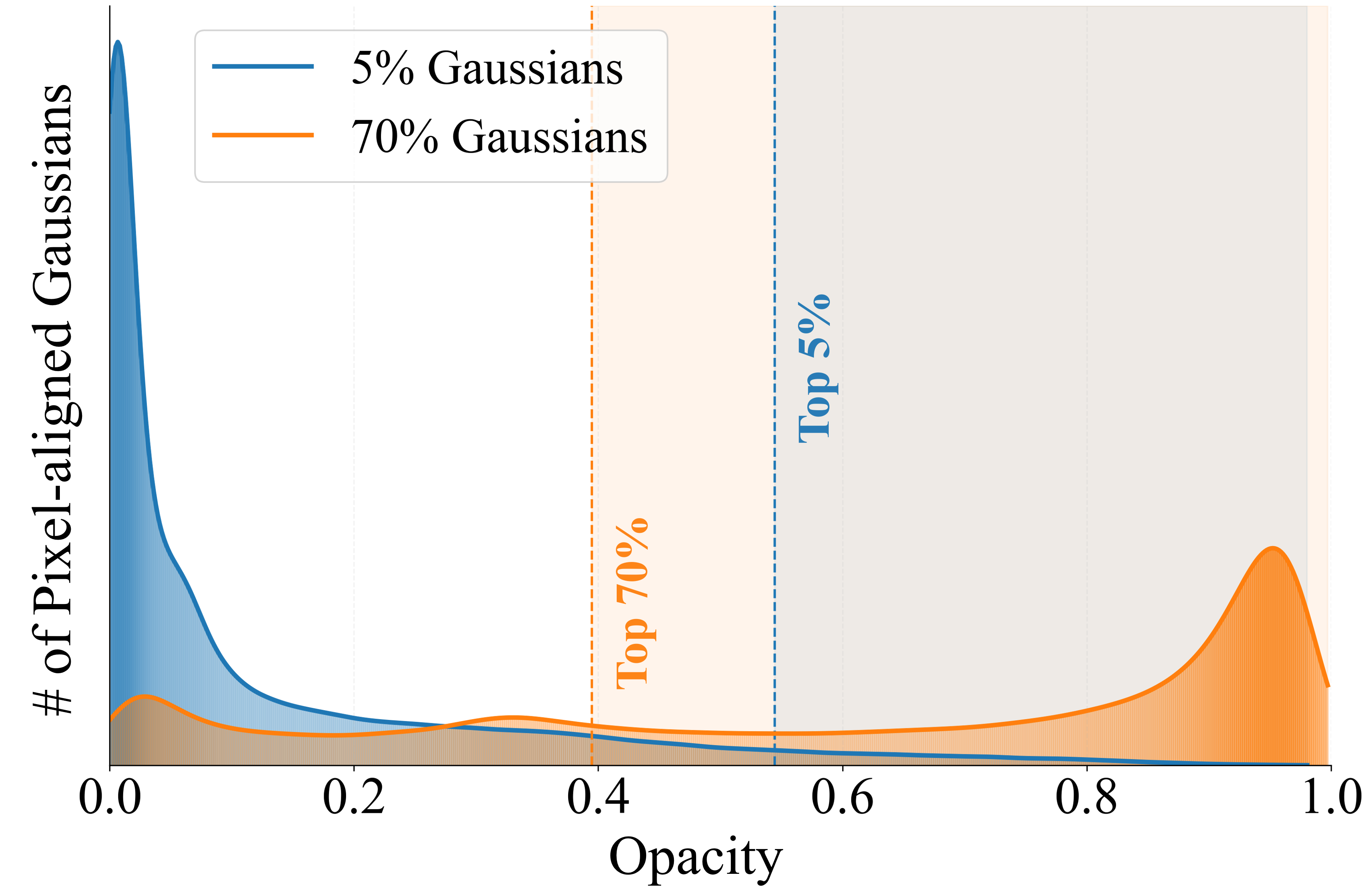}
    \vspace{-0.3cm}
    \caption{\textbf{Opacity distributions of pixel-aligned Gaussians across target primitive counts $K$.} The blue and orange curves show the distributions when the target primitive count is set to 5\% and 70\% of the total number of pixel-aligned Gaussians, respectively. EcoSplat adaptively predicts Gaussian opacities based on the primitive budget and selects the top-$K$ Gaussians.}
    \label{fig:opacity}
\end{figure}

\noindent \textbf{Importance-Aware Opacity Analysis.}
In Fig.~\ref{fig:opacity}, we visualize the opacity distributions predicted by EcoSplat for varying target primitive counts $K$ on a given scene. When $K$ is large (`70\% Gaussians', orange), the model retains a substantial proportion of high-opacity primitives that capture fine geometric and appearance details.
In contrast, when $K$ is reduced to 5\% (blue), EcoSplat effectively suppresses the opacities of low-importance primitives, pushing most of them toward the low-opacity region. Only a compact, highly informative subset maintains high opacity, corresponding to the most structurally and photometrically critical elements of the scene. This behavior illustrates that EcoSplat learns a robust importance ranking and dynamically adjusts Gaussian visibility according to the target primitive counts $K$.

\begin{table}[t]
\centering
\setlength\tabcolsep{3pt}
\renewcommand{\arraystretch}{1.1}
\scalebox{0.8}{
\begin{tabular}{l | c c c | c c c}
\toprule
\multicolumn{1}{c|}{\multirow{2}{*}{Variants}} & \multicolumn{3}{c|}{5\% Gaussians} & \multicolumn{3}{c}{40\% Gaussians} \\
\cline{2-7}
& PSNR$\uparrow$ & SSIM$\uparrow$ & LPIPS$\downarrow$ & PSNR$\uparrow$ & SSIM$\uparrow$ & LPIPS$\downarrow$ \\
\midrule
w/o PGT & 22.93 & 0.778 & 0.221 & 23.70 & 0.806 & 0.184 \\
w/o IGF & 6.45 & 0.107 & 0.651 & 14.02 &0.586 & 0.341 \\
w/o $\mathcal{L}_\text{io}$ & 20.58 & 0.724 & 0.289 & 23.81 & 0.815 & 0.167 \\
w/o PLGC & 21.49 & 0.725 & 0.280 & 23.84 & 0.799 & 0.193 \\
\textbf{Ours} & 24.72 & 0.822 & 0.183 & 25.11 & 0.835 & 0.164 \\
\bottomrule
\end{tabular}
}
\vspace{-0.15cm}
\caption{\textbf{Ablation study on RE10K~\cite{zhou2018stereo} with 24 input views.}}
\label{tab:ablation}
\vspace{-0.25cm}
\end{table}

\section{Conclusion}
We present {EcoSplat}, the \textbf{first} feed-forward 3D Gaussian Splatting framework with \textbf{explicit control over the number of output Gaussian primitives} from multi-view images. EcoSplat enables flexible rendering on a wide range of end-user devices while maintaining high rendering quality. We validate the effectiveness of EcoSplat across extensive input-view settings and under various numbers of target primitives. Our experiments demonstrate that EcoSplat significantly outperforms SOTA methods using fewer primitives, and provides the freedom to dynamically control the primitive counts, offering an optimal trade-off between efficiency and rendering fidelity.


\clearpage
\setcounter{page}{1}
\maketitlesupplementary

\appendix

\section{Additional Experimental Results}
 
\subsection{Efficiency Comparisons}
In Table~\ref{table:latency}, we present a detailed efficiency comparison of all feed-forward 3DGS methods evaluated in Table~\ref{table:re10k_multi_view}. The reported metrics include `Recon. Latency (s)', `Render FPS', `\# GS', and `Storage (MB)', which respectively correspond to the time required to generate 3D Gaussians from input views, the rendering speed obtained from rasterizing the predicted 3D Gaussians, the number of output 3D Gaussians, and their memory footprint.
As shown, EcoSplat substantially improves both rendering FPS and storage efficiency by predicting far fewer Gaussians. Moreover, EcoSplat achieves the fastest reconstruction time among all feed-forward methods. Although NoPoSplat~\cite{ye2024no} uses a similar MASt3R~\cite{mast3r_eccv24} backbone, its architecture requires performing cross-attention between the first input view and each remaining view independently, resulting in substantial computational overhead. Following SPFSplat~\cite{huang2025no}, EcoSplat adopts a global attention mechanism that aggregates feature information from all views. Furthermore, whereas SPFSplat predicts Gaussian parameters sequentially across views, EcoSplat predicts all per-view Gaussian primitives in parallel. This parallel implementation significantly reduces reconstruction latency, making EcoSplat the most efficient method overall.
\vspace{-0.2cm}

\subsection{Additional NVS Results under Controlled Number of Primitives} 
We further evaluate all methods under a more challenging test setup. Instead of randomly sampling input and target views, we require each target view to be at least 20 frames away from its nearest input view, which forces the models to synthesize substantially more extreme novel viewpoints.
We additionally include per-scene optimization methods that support controllable primitive counts, such as 3DGS-MCMC~\cite{kheradmand20243d}, PUP 3D-GS~\cite{HansonTuPUP3DGS}, and Lightgaussian~\cite{fan2023lightgaussian}, to compare them with feed-forward approaches.
For 3DGS-MCMC~\cite{kheradmand20243d}, we train the model for 30K iterations.
For PUP 3D-GS~\cite{HansonTuPUP3DGS} and Lightgaussian~\cite{fan2023lightgaussian}, we first optimize the 3DGS representation for 10K iterations, prune the primitives using their respective strategies, and then resume training for an additional 10K iterations.
These per-scene optimization methods require approximately 10 minutes of reconstruction time.
Moreover, we finetune the 3D Gaussian parameters predicted by the cascaded baselines that combine SPFSplat~\cite{huang2025no} with post-pruning methods~\cite{fan2023lightgaussian, HansonTuPUP3DGS}, and report results after 1K and 10K finetuning iterations, respectively.
The results demonstrate that our feed-forward EcoSplat significantly outperforms all competing methods, including per-scene optimization approaches, by large margins. EcoSplat exhibits significantly less overfitting to the input views and demonstrates greater robustness under large viewpoint shifts than the per-scene optimization approaches~\cite{kheradmand20243d, HansonTuPUP3DGS, fan2023lightgaussian}. Although the cascade baselines (SPFSplat~\cite{huang2025no} + Lightgaussian~\cite{fan2023lightgaussian}, SPFSplat~\cite{huang2025no} + PUP 3D-GS~\cite{HansonTuPUP3DGS}) show improvements after per-scene post-optimization, they still fall behind our EcoSplat without any further Gaussian finetuning. Note that per-scene post-optimization for these cascade baselines requires up to 1 minute of finetuning, adding substantial overhead to the feed-forward pipeline.

\begin{table}
\begin{center}
\setlength\tabcolsep{6pt} 
\renewcommand{\arraystretch}{1.1}
\scalebox{0.8}{
\begin{tabular}{ l | c c c c}
\toprule
\multicolumn{1}{c|}{\multirow{2}{*}{Methods}} & Recon. & Render & \multirow{2}{*}{\# GS$\downarrow$} & Storage\\
& Latency (s)$\downarrow$ &  FPS$\uparrow$ & &  (MB)$\downarrow$ \\
\midrule
MVSplat~\cite{chen2024mvsplat} & 1.41 & 679 & 1573K & 534 \\
DepthSplat~\cite{xu2025depthsplat}  & \second{0.56} & 683 & 1573K & 534 \\
NoPoSplat~\cite{ye2024no}  & 2.65 & 715 & 1573K & 534 \\
SPFSplat~\cite{huang2025no} &  0.61 & 710 & 1573K & 534 \\
GGN~\cite{zhang2024gaussian}  &  1.59 & \second{987} & \second{512K} & \second{174} \\
AnySplat~\cite{jiang2025anysplat} &  1.63 & 780 & 1259K & 428 \\
WorldMirror~\cite{liu2025worldmirror} & 0.64 & 830 & 1020K & 346 \\ 
\midrule
Ours$_{5\%}$ & \best{0.52} & \best{1042} &  \best{78K}& \best{27} \\
Ours$_{40\%}$  & \best{0.52} & 977 &  {629K}& 214 \\
\bottomrule
\end{tabular}
}
\vspace{-0.1cm}
\caption{\textbf{Efficiency comparison.} We report the efficiency metrics of the existing methods under the 24-view setting. All evaluations are conducted on a single NVIDIA A100 GPU. To ensure a fair comparison of rendering FPS, we first generate and store the 3D Gaussian primitives for every method, and then render the target views using the same rasterization backend across all methods.}
\label{table:latency}
\vspace{-1cm}
\end{center}
\end{table}

\begin{table*}
\begin{center}
\setlength\tabcolsep{6pt} 
\renewcommand{\arraystretch}{1.1}
\scalebox{0.75}{
\begin{tabular}{ l | c c c | c c c | c c c | c c c}
\toprule
\multicolumn{1}{c|}{\multirow{2}{*}{Methods}} & \multicolumn{3}{c|}{5\% Gaussians}  & \multicolumn{3}{c|}{10\% Gaussians}  & \multicolumn{3}{c|}{40\% Gaussians} & \multicolumn{3}{c}{70\% Gaussians} \\
\cline{2-13}
& PSNR$\uparrow$ & SSIM$\uparrow$ & LPIPS$\downarrow$ & PSNR$\uparrow$ & SSIM$\uparrow$ & LPIPS$\downarrow$ & PSNR$\uparrow$ & SSIM$\uparrow$ & LPIPS$\downarrow$ & PSNR$\uparrow$ & SSIM$\uparrow$ & LPIPS$\downarrow$ \\
\midrule
3DGS-MCMC~\cite{kheradmand20243d} & 20.34 & 0.698 & 0.316 & 20.50 & 0.710 & 0.304 & 20.28 & 0.715 & 0.313 & 19.97 & 0.710 & 0.329 \\
PUP 3D-GS~\cite{HansonTuPUP3DGS} & 15.72 & 0.600 & 0.503 & 16.69 & 0.635 & 0.452 & 17.01 & 0.653 & 0.422 & 17.30 & 0.658& 0.417 \\

Lightgaussian~\cite{fan2023lightgaussian} & 15.82 & 0.605 & 0.477 & 16.44 & 0.631 & 0.459 & 17.57 & 0.663 & 0.420 & 17.43 & 0.658& 0.415 \\

SPFSplat~\cite{huang2025no} +~\cite{fan2023lightgaussian} + 1K iter. & 22.08 & 0.774 & 0.314 & 22.79 & 0.799 & 0.272 & 23.92 & 0.830 & 0.218 & 24.38 & 0.838 & 0.206\\
SPFSplat~\cite{huang2025no} +~\cite{HansonTuPUP3DGS} + 1K iter. & 22.67 & 0.803 & 0.259 & 23.39 & 0.821 & 0.229 & \second{24.13} & \second{0.835} & \second{0.210} & \second{24.50} & \second{0.840} & {0.205}\\
SPFSplat~\cite{huang2025no} +~\cite{fan2023lightgaussian} + 10K iter. & 22.08 & 0.774 & 0.314 & 22.79 & 0.799 & 0.272 & 24.07 & 0.807 & 0.232 & 24.06 & 0.804 & 0.244\\
SPFSplat~\cite{huang2025no} +~\cite{HansonTuPUP3DGS} + 10K iter. & \second{23.73} & \second{0.819} & \second{0.224} & \second{23.99} & \second{0.822} & \second{0.219} & 24.05 & 0.807 & 0.239 & 24.07 & 0.802 & 0.245\\
\midrule
AnySplat~\cite{jiang2025anysplat} & 7.92 & 0.268 & 0.574 & 10.97 & 0.440 & 0.469 & 18.73 & 0.676 & 0.253 & 20.13 & 0.714 & \second{0.189} \\
WorldMirror~\cite{liu2025worldmirror} & 7.31 & 0.222 & 0.627 & 9.74 & 0.381 & 0.526 & 18.37 & 0.683 & 0.287 & 20.55 & 0.736 & 0.203 \\
MVSplat~\cite{chen2024mvsplat} +~\cite{fan2023lightgaussian} & 4.75 & 0.053 & 0.727 & 5.32 & 0.098 & 0.697 & 9.32& 0.368 & 0.509 & 10.84 & 0.430 & 0.583 \\
MVSplat~\cite{chen2024mvsplat} +~\cite{HansonTuPUP3DGS} & 4.86 & 0.065 & 0.688 & 5.65 & 0.114 & 0.663 & 9.89 & 0.386 & 0.491 & 10.84 & 0.429 & 0.455 \\
SPFSplat~\cite{huang2025no} +~\cite{fan2023lightgaussian} & 6.64 & 0.243 & 0.647 & 7.34 & 0.285 & 0.612 & 10.51 & 0.519 & 0.415 & 13.67 & 0.669 & 0.274 \\
SPFSplat~\cite{huang2025no} +~\cite{HansonTuPUP3DGS}& 5.31 & 0.136 & 0.610 & 6.37 & 0.224 & 0.564 & 11.54 & 0.556 & 0.367 & 14.75 & 0.690 & 0.252\\
\midrule
EcoSplat (Ours) & \best{24.80} & \best{0.843} & \best{ 0.157} & \best{24.87} & \best{0.845} & \best{0.156} & \best{25.02} & \best{0.847} & \best{0.151} & \best{24.90} & \best{0.845} & \best{0.151} \\
\bottomrule
\end{tabular}
}

\vspace{-0.2cm}
\caption{\textbf{Addition quantitative comparison of NVS under controlled numbers of primitives on the RE10K dataset~\cite{zhou2018stereo} with 24 input views.} 
We evaluate the baselines on more challenging test views than those used in Table~\ref{table:primitive_ratio}. Moreover, we further compare EcoSplat against additional baselines, including per-scene optimization methods and post-optimization variants of the cascaded methods. `+ 1K iter.' and `+ 10K iter.' indicate that we further optimize the parameters of the output 3D Gaussians from each cascaded method for an additional 1K and 10K iterations, respectively.}
\label{table:finetune}
\vspace{-0.4cm}
\end{center}
\end{table*}

\subsection{Ablation Study on Architectural Design}
\begin{figure*}[ht]
    \centering    \includegraphics[scale=0.12]{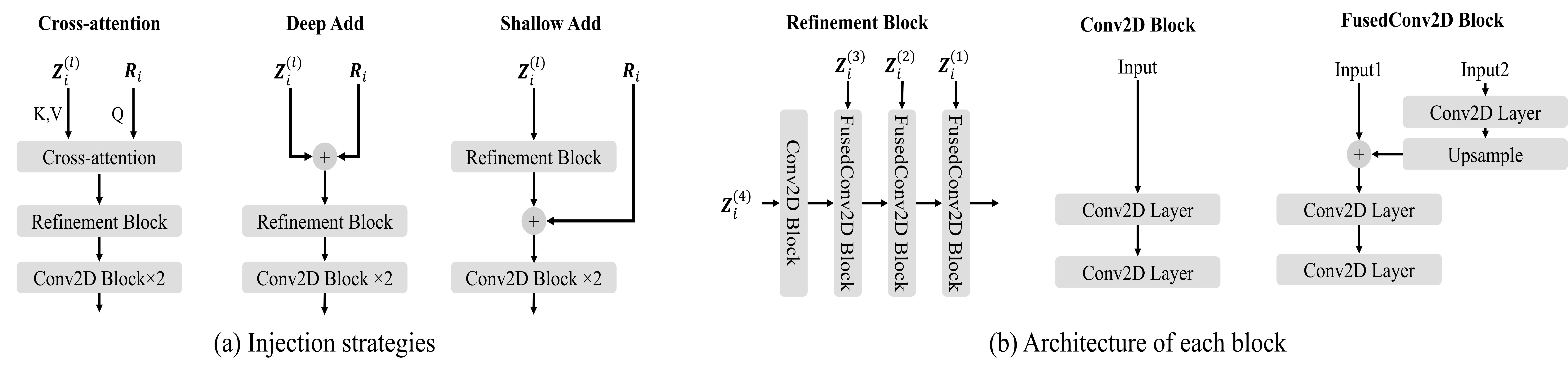}
    \vspace{-0.5cm}
    \caption{\textbf{Importance embedding injection strategies.}}
    \label{fig:ablation_architecture}
\end{figure*}

In Table~\ref{tab:ablation_architecture}, we ablate the injection strategies of our learnable importance embedding $\bm{R}_i$ in Eq.~\ref{eq:stage2_GS_params}, with the corresponding designs illustrated in Fig.~\ref{fig:ablation_architecture}-(a).
In the `Cross-attention' variant, each $\bm{R}_i$ interacts with its corresponding $\bm{Z}_i^{(\ell)}$ through cross-attention.
The `Deep Add' variant replaces this cross-attention operation with a simple additive fusion of $\bm{R}_i$ and $\bm{Z}_i^{(\ell)}$.
The fused features are then passed to a residual convolutional refinement block.
Our final design, the `Shallow Add' variant, fuses $\bm{R}_i$ with the output of the residual convolutional refinement block applied to $\{\bm{Z}_i^{(\ell)}\}_{\ell=1}^{m}$ by simple addition.
This refinement block's architecture is shown in Fig.~\ref{fig:ablation_architecture}-(b).
Among all variants, the `Shallow Add' variant achieves the best balance of effectiveness and stability.

\begin{table}[ht]
\centering
\setlength\tabcolsep{3pt}
\renewcommand{\arraystretch}{1.1}
\scalebox{0.7}{
\begin{tabular}{l | c c c | c c c}
\toprule
\multicolumn{1}{c|}{\multirow{2}{*}{Variants}} & \multicolumn{3}{c|}{5\% Gaussians} & \multicolumn{3}{c}{40\% Gaussians} \\
\cline{2-7}
& PSNR$\uparrow$ & SSIM$\uparrow$ & LPIPS$\downarrow$ & PSNR$\uparrow$ & SSIM$\uparrow$ & LPIPS$\downarrow$ \\
\midrule
Cross-attention & \second{24.71} & \best{0.828} & \best{0.179} & \second{24.90} & \second{0.831} &\second{ 0.168} \\
Deep Add & 24.18 & 0.814 & 0.196 & 24.62 &0.824& 0.178 \\
\textbf{Ours} & \best{24.72} & \second{0.822} & \second{0.183} & \best{25.11} & \best{0.835} & \best{0.164} \\
\bottomrule
\end{tabular}
}
\vspace{-0.15cm}
\caption{\textbf{Ablation study of the learnable importance embedding injection on RE10K~\cite{zhou2018stereo} with 24 input views.} The `Shallow Add' injection strategy achieves superior performance compared to the `Cross-attention' and `Deep Add' variants.}
\label{tab:ablation_architecture}
\vspace{-0.25cm}
\end{table}

\subsection{Additional Cross-Dataset Generalization NVS}
We provide extended quantitative results for the 16-view and 20-view settings in Table~\ref{table:acid_multi_view_16_20}, complementing the evaluations reported in Table~\ref{table:acid_multi_view}.
\begin{table}
\begin{center}
\setlength\tabcolsep{6pt} 
\renewcommand{\arraystretch}{1.1}
\scalebox{0.56}{
\begin{tabular}{ l | c c c c | c c c c }
\toprule
\multicolumn{1}{c|}{\multirow{2}{*}{Methods}} & \multicolumn{4}{c|}{16 Views}  & \multicolumn{4}{c}{20 Views}   \\
\cline{2-9}
& PSNR$\uparrow$ & SSIM$\uparrow$ & LPIPS$\downarrow$ & \# GS$\downarrow$ & PSNR$\uparrow$ & SSIM$\uparrow$ & LPIPS$\downarrow$ & \# GS$\downarrow$  \\
\midrule
MVSplat~\cite{chen2024mvsplat} & 18.19 & 0.502 & 0.376 & 1048K & 18.12 & 0.500 & 0.379 & 1310K  \\
DepthSplat~\cite{xu2025depthsplat} & 20.41 & \second{0.713} & 0.268 & 1048K & 19.92 & 0.694 & 0.286 & 1310K  \\
NoPoSplat~\cite{ye2024no} & 16.77 & 0.500 & 0.409 & 1048K & 16.84 & 0.497 & 0.413 & 1310K\\
SPFSplat~\cite{huang2025no} & \best{24.58} & \best{0.725} & \best{0.218} & 1048K & \best{24.49} & \best{0.722} & \best{0.221} & 1310K \\
GGN~\cite{zhang2024gaussian} & 18.46 & 0.564 & 0.380 & \second{391K} & 18.32 & 0.560 & 0.384 & \second{436K}  \\
AnySplat~\cite{jiang2025anysplat} & 21.89 & 0.615 & 0.262 & 861K & 22.05 & 0.622 & 0.256 & 1050K  \\
WorldMirror~\cite{liu2025worldmirror} & 22.15 & 0.646 & 0.275 & 806K & 22.20 & 0.650 & 0.275 & 978K \\ 
\midrule
Ours$_{5\%}$ & 23.92 &0.689 & 0.282 &  \best{52K} & 23.95 & 0.690 & 0.282 & \best{78K}  \\
Ours$_{40\%}$ & \second{24.17}& {0.701} & \second{0.250} &  {419K} & \second{24.12} & \second{0.700} & \second{0.252} &  {524K} \\
\bottomrule
\end{tabular}
}
\vspace{-0.1cm}
\caption{\textbf{Cross-dataset generalization with 16 and 20 input views.} We train all methods on the RE10K dataset~\cite{zhou2018stereo} and evaluate their zero-shot performance on the ACID dataset~\cite{infinite_nature_2020}.}
\label{table:acid_multi_view_16_20}
\vspace{-1cm}
\end{center}
\end{table}

\subsection{Long-LRM Results}
We acknowledge the concurrent work Long-LRM~\cite{ziwen2025long}, as discussed in the Related Works section (Sec.~\ref{sec:related_work}). Despite our extensive efforts to include this model in our comparison, as of 20 November 2025, no official training configuration or released code for the RE10K dataset has been made publicly available. We attempted to train Long-LRM by reproducing all available details and using the checkpoint provided by the authors. However, the absence of an official RE10K configuration and incomplete training specifications led to repeated training failures, and the resulting outputs were not reliable enough for a fair comparison. Therefore, although we made every reasonable effort to reproduce Long-LRM with the currently available public resources, we were unable to obtain valid experimental results. Once the official configuration becomes available, we will gladly include a full experimental comparison.

\section{Limitation and Future Work}
Similar to prior feed-forward 3DGS frameworks~\cite{xu2025depthsplat, chen2024mvsplat, charatan2024pixelsplat, ye2024no, huang2025no, jiang2025anysplat, liu2025worldmirror}, EcoSplat is currently designed for NVS of \textit{static} scenes from multi-view images. Therefore, our model does not yet handle object deformation or dynamic scene reconstruction. Nevertheless, recent progress in \textit{feed-forward} dynamic scene reconstruction~\cite{xiao2025spatialtrackerv2, wang20254realvideov2fusedviewtimeattention} have proposed a promising extension that finetunes the global cross-attention layers to learn temporal consistency in dynamic scenes. Inspired by this, our framework could be extended to handle dynamic scenes and jointly predict global static primitives and per-frame dynamic primitives.






{
    \small
    \bibliographystyle{ieeenat_fullname}
    \bibliography{main}
}


\end{document}